%% file: Formatting-Instructions-LaTeX-2026.tex
\documentclass[letterpaper]{article} % DO NOT CHANGE THIS
\usepackage{aaai2026}  % DO NOT CHANGE THIS
\usepackage{times}  % DO NOT CHANGE THIS
\usepackage{helvet}  % DO NOT CHANGE THIS
\usepackage{courier}  % DO NOT CHANGE THIS
\usepackage[hyphens]{url}  % DO NOT CHANGE THIS
\usepackage{graphicx} % DO NOT CHANGE THIS
\usepackage{natbib}  % DO NOT CHANGE THIS AND DO NOT ADD ANY OPTIONS TO IT
\usepackage{caption} % DO NOT CHANGE THIS AND DO NOT ADD ANY OPTIONS TO IT
\usepackage{algorithm}
\usepackage{algorithmic}
\usepackage{amsmath}
\usepackage{amssymb}
\usepackage{booktabs}
\usepackage{multirow}
\usepackage{graphicx}
\usepackage{subcaption}
\usepackage{bm}

\usepackage{newfloat}
\usepackage{listings}
\DeclareCaptionStyle{ruled}{labelfont=normalfont,labelsep=colon,strut=off} % DO NOT CHANGE THIS
\floatstyle{ruled}
\newfloat{listing}{tb}{lst}{}
\floatname{listing}{Listing}
\title{FIND: A Simple Yet Effective Baseline for Diffusion-Generated Image Detection}

\author {
    Jie Li\textsuperscript{\rm 1},
    Yingying Feng\textsuperscript{\rm 2},
    Chi Xie\textsuperscript{\rm 3},
    Jie Hu\textsuperscript{\rm 4},
    Lei Tan\textsuperscript{\rm 4}\thanks{Corresponding Author.},
    Jiayi Ji\textsuperscript{\rm 1,\rm 4}
}
\affiliations {
    \textsuperscript{\rm 1}Xiamen University \\
    \textsuperscript{\rm 2}Northeastern University \\
    \textsuperscript{\rm 3}Tongji University \\
    \textsuperscript{\rm 4}National University of Singapore\\
    lijie.32@outlook.com, lei.tan@nus.edu.sg
}

\begin{document}

\maketitle

\input{sec/0_abstract}    
\input{sec/1_intro}
\input{sec/2_related_work}
\input{sec/3_method}
\input{sec/4_exp}
\input{sec/5_conclusion}

\section{Acknowledgements}
This work is supported by the China Postdoctoral Science Foundation (No.2025M771514).

\bibliography{aaai2026}

\end{document}

%% file: sec/0_abstract.tex
\begin{abstract}
The remarkable realism of images generated by diffusion models poses critical detection challenges. Current methods utilize reconstruction error as a discriminative feature, exploiting the observation that real images exhibit higher reconstruction errors when processed through diffusion models. However, these approaches require costly reconstruction computations and depend on specific diffusion models, making their performance highly model-dependent. We identify a fundamental difference: real images are more difficult to fit with Gaussian distributions compared to synthetic ones. In this paper, we propose Forgery Identification via Noise Disturbance (FIND),  a novel method that requires only a simple binary classifier. It eliminates reconstruction by directly targeting the core distributional difference between real and synthetic images. Our key operation is to add Gaussian noise to real images during training and label these noisy versions as synthetic. This step allows the classifier to focus on the statistical patterns that distinguish real from synthetic images. We theoretically prove that the noise-augmented real images resemble diffusion-generated images in their ease of Gaussian fitting. Furthermore, simply by adding noise, they still retain visual similarity to the original images, highlighting the most discriminative distribution-related features. The proposed FIND improves performance by 11.7\% on the GenImage benchmark while running 126$\times$ faster than existing methods. With no need for auxiliary diffusion models and reconstruction, it offers a practical, efficient, and generalizable way to detect diffusion-generated content.
 % in real-world applications.
% \footnote{The source codes will be available upon accepted.}.
% }

\end{abstract}

%%% Local Variables:
%%% mode: LaTeX
%%% TeX-master: "../main"
%%% End:

%% file: sec/1_intro.tex
\section{Introduction}
\label{sec:intro}

\begin{figure}[t]
	\centering
	\includegraphics[width=0.85\columnwidth]{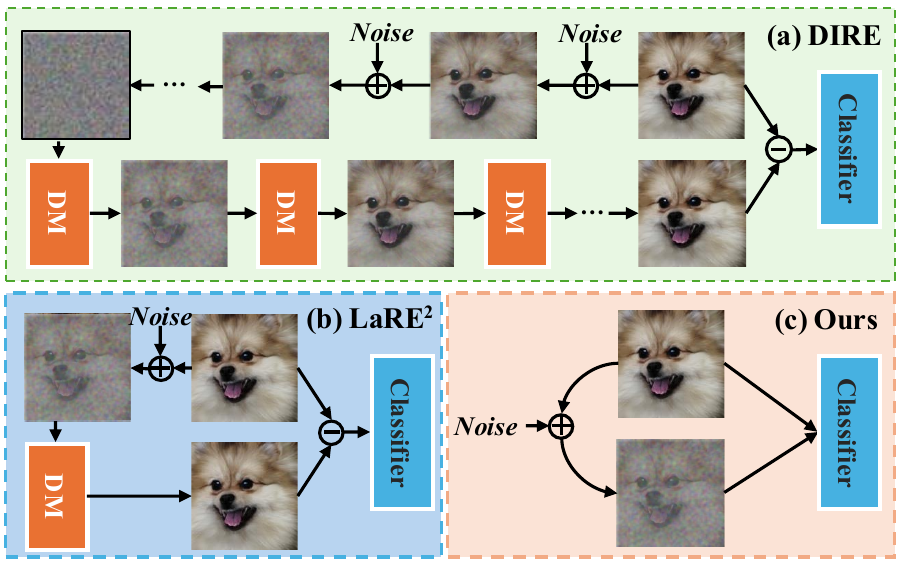}
	\caption{Comparison between FIND and previous noise-based methods.
    (a) DIRE utilizes complete reconstruction by adding noise and then denoising it with a Diffusion Model (DM) in multiple steps;
(b) LaRE$^2$ compresses the noise addition and denoising steps into a single step within the latent space;
(c) Our method only involves adding noise without the denoising process.
Compared with previous methods, FIND avoids explicit reconstruction and omits the dependence on DM.
    }
	\label{fig:motivation}
	% \vspace{-1.5em}
\end{figure}

The rapid advancement of diffusion models~\cite{2105.05233v3,ho2020denoising,nichol2021glide,rombach2022high} has revolutionized the field of generative artificial intelligence, enabling the synthesis of highly realistic images that are often indistinguishable from real photographs. 
While this technological breakthrough has opened new possibilities in creative domains~\cite{zhou2024storydiffusionconsistentselfattentionlongrange,pan2022synthesizing,he2025survey}, it has also introduced significant challenges in ensuring the trustworthiness of digital media~\cite{juefei2022countering,carlini2023extractingtrainingdatadiffusion}. 
The ability to generate convincing synthetic images raises critical concerns regarding misinformation, intellectual property rights, and digital security.
As a result, the detection of generated images has become urgent and non-trivial.

Current mainstream approaches~\cite{wang2023dire,2403.17465v3,ma2023exposing} to detecting diffusion-generated images primarily rely on sophisticated reconstruction-based methods, which utilize an additional diffusion model to assess reconstruction errors. 
% %
Based on \emph{reconstruction assumption} that diffusion-generated images are more easily reconstructed by a diffusion model compared with real images,
DIRE~\cite{wang2023dire} uses the reconstruction error as image representations and employ a binary classifier on them.
Following DIRE, SeDID~\cite{ma2023exposing} further exploits intermediate steps during the reconstruction process.
Among these methods, an extra diffusion model, serving as the reconstruction model, needs to infer multiple steps to achieve satisfactory results as shown in Fig.\,\ref{fig:motivation}(a), which leads to low efficiency.
To improve efficiency, LaRE~\cite{2403.17465v3} introduces one-step solutions based on reconstruction from the latent space depicted in Fig.\,\ref{fig:motivation}(b).
These methods, while effective, suffer from two major limitations: poor generalizability across various image generation models and prohibitive computational resource requirements.
The generalizability issue arises from the specific diffusion model used to calculate the reconstruction errors, while the computational overhead stems from the need to process each image through a complex reconstruction pipeline.

The limitations of existing methods motivate us to consider the detection method without an explicit reconstruction process along with an extra reconstruction model.
We start by hypothesizing that the discriminative features derived from the current reconstruction process fundamentally stem from the differing abilities of Gaussian distributions to approximate image statistics, which can be easily derived from the \emph{reconstruction assumption}. Specifically, real images are more difficult to fit with Gaussian distributions compared to their synthetic counterparts.
This hypothesis provides a new perspective on understanding the nature of the differences between real and synthetic images.
However, directly using a classifier without prior knowledge to identify this crucial discriminative feature is extremely challenging. 
The image distribution contains multiple dimensions and various features like visual elements, which can interfere with the classifier's ability to focus on the key discriminative information. 
Previous methods tackle these by directly providing reconstruction error as a discriminative feature. This approach eliminates visual and other interference factors as much as possible through the process of subtracting reconstructed images from original ones but inevitably introduces additional computational overhead.

In this paper, we propose a novel method named \textbf{F}orgery \textbf{I}dentification via \textbf{N}oise \textbf{D}isturbance (FIND) to provide the discriminative feature implicitly, with solely the binary classifier as shown in Fig.\,\ref{fig:motivation}(c).
% In this paper, we propose a novel method named FIND to provide the discriminative feature implicitly.
%
Specifically, during the training phase,
we add random Gaussian noise to real images and label these noisy images as synthetic images. 
This operation serves two purposes. 
Firstly, similar to synthetic images, real images with added noise are more easily fitted by Gaussian distributions compared to the original real images, which helps to emphasize the discriminative feature as we hypothesized before.
This point is theoretically proven in Section~\ref{sec:Hypothesis}. Secondly, the noisy images retain a certain visual similarity to the original real images. This similarity allows us to eliminate other interfering features in the image distribution, enabling the classifier to concentrate more on the discriminative feature.
We use real images, noisy real images, and synthetic images to train a basic binary classifier.
During the inference phase, the images are directly input into the binary classifier to get the prediction.
Our proposed method offers several advantages over existing approaches. 
Most notably, it eliminates the need for a specific reconstruction model and the reconstruction process, which significantly improves the generalizability of the detection method and reduces both the training and inference time.
Specifically, our method demonstrates superior performance on the large-scale GenImage benchmark~\cite{2306.08571v2}, achieving an 11.7\% performance improvement compared to existing methods. 
And, due to its simple architecture, our method is $126 \times$ faster than the state-of-the-art methods, making it more practical for real-world applications.

In conclusion, this paper presents a simple yet effective method for detecting images generated by diffusion models. It not only provides a new theoretical understanding of the discriminative features in the reconstruction process, but also demonstrates that noised real images can serve as boundary samples, maintaining content consistency with real images while exhibiting distribution patterns similar to synthetic images. Based on this insight, we offer a more practical baseline for future research. 
The extensive experiments on the large-scale benchmark demonstrate the effectiveness and efficiency of our method.
And our method will provide a brand-new baseline for this field.

\input{sub_tex/figures/fig2_3}

%%% Local Variables:
%%% mode: LaTeX
%%% TeX-master: "../anonymous-submission-latex-2026.tex"
%%% End:

%% file: sub_tex/figures/fig2_3.tex
\begin{figure*}[t]
\centering
\begin{minipage}[t]{0.48\textwidth}
	\includegraphics[width=\linewidth]{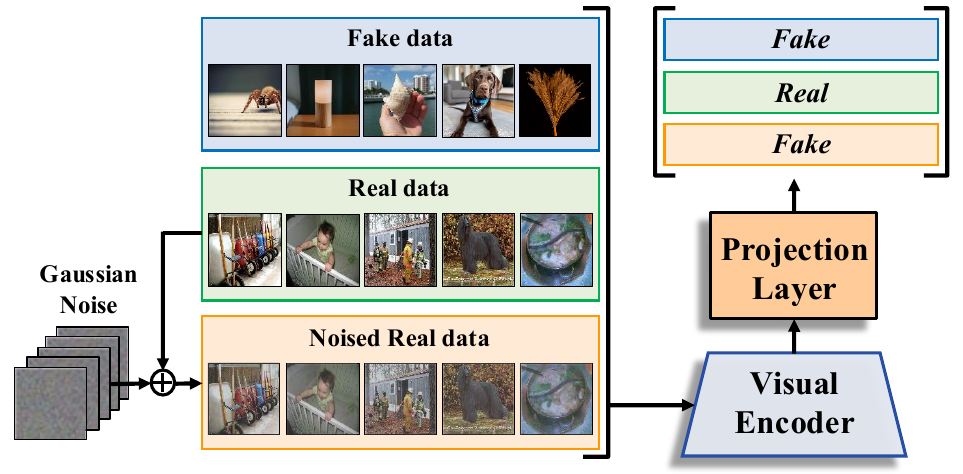}
    \caption{The training framework of FIND. Gaussian noise is added to real images, and these perturbed versions are labeled as synthetic within the training process. This enables FIND to mitigate the reconstruction model dependency, and learn the core distributional differences between real and synthetic images, resulting in a superior generalization ability.}
	\label{fig:ci_vis}
\end{minipage}
\hfill
\begin{minipage}[t]{0.48\textwidth}
\includegraphics[width=\linewidth]{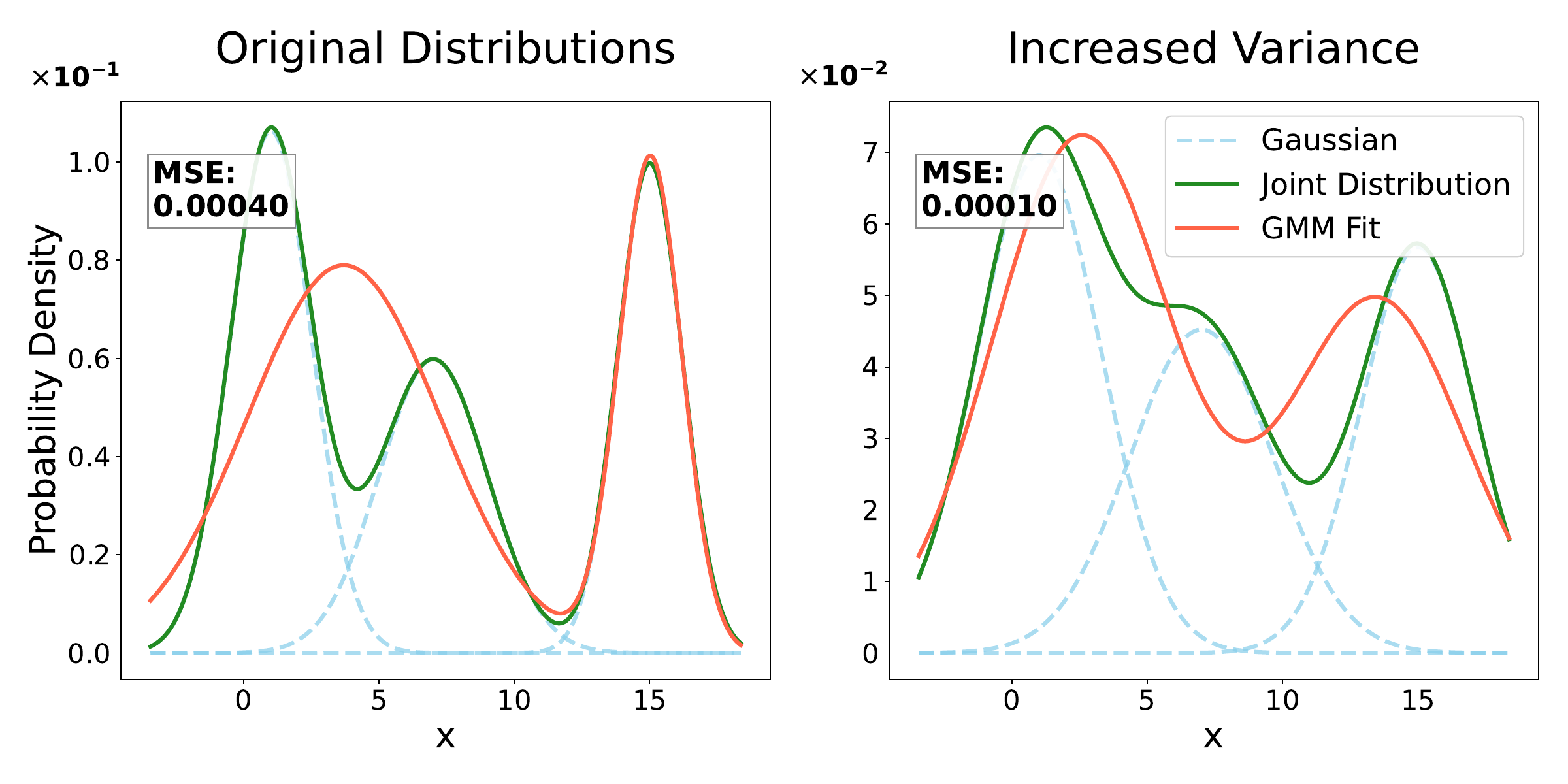}
\caption{\label{fig:gmm} 
Visual illustration of Gaussian distribution with increased variances. 
% Left panel: The green line represents the joint distribution formed by three Gaussian distributions, and the red line shows the fitting results using a two-component Gaussian mixture model.
% Right panel: After increasing the variances of the three Gaussian distributions, the joint distribution becomes smoother, with the two left-hand Gaussian distributions nearly merging. The mean squared error of the fitting drops from 0.00039 to 0.00010 for the distribution in the right-hand figure, which means distributions with increased variance are easier to fit.
\textbf{Left panel}: A joint distribution (green) from three Gaussians, with its two-component Gaussian mixture model fit (red).
\textbf{Right panel}: Increasing the variances smooths the joint distribution, causing the two left-hand Gaussians to nearly merge. The drop of fitting MSE indicates that distributions with increased variance are easier to fit.
}
\end{minipage}
\end{figure*}
%%% Local Variables:
%%% mode: LaTeX
%%% TeX-master: "../../main"
%%% End:

%% file: sec/2_related_work.tex
\section{Related Work}
\label{sec:related_work}

\subsection{Diffusion Models for Image Generation}
Diffusion models have emerged as a transformative paradigm for high-fidelity image generation, building on foundational principles from non-equilibrium thermodynamics. 
Early breakthroughs like Denoising Diffusion Probabilistic Models~\cite{ho2020denoising} (DDPMs) established the framework of iterative denoising, achieving performance competitive with leading GAN-based approaches~\cite{Karras2019stylegan2}. 
Subsequent research has focused on enhancing their practicality and versatility, with innovations spanning accelerated sampling algorithms~\cite{nichol2021improved,songdenoising,liupseudo}, architectural refinements~\cite{2105.05233v3,rombach2022high}, and novel conditioning mechanisms for tasks like text-to-image synthesis~\cite{rombach2022high,gu2022vectorvqdm,zhang2023adding}. For instance, latent diffusion models~\cite{rombach2022high} significantly improved computational efficiency with compressed representations, while cross-attention mechanisms enabled fine-grained multimodal control~\cite{zhang2023adding,ruiz2023dreambooth}. Advancements in deterministic sampling~\cite{songdenoising} and pseudo-numerical methods~\cite{liupseudo} reduced inference steps without compromising quality, bridging the gap between probabilistic and implicit modeling. Beyond generation, diffusion models have been adapted for diverse applications, from image editing~\cite{nichol2021glide,hertzprompt}, inpainting~\cite{lugmayr2022repaint}, to detection of synthetic content~\cite{juefei2022countering}, reflecting their broad impact. These developments, coupled with community-driven efforts like Stable Diffusion~\cite{rombach2022high}, underscore their scalability and adaptability, positioning diffusion models as a cornerstone of modern generative AI.

\subsection{Generated Image Detection}
The rapid evolution of generative models has spurred significant advancements in synthetic image detection, with methods evolving alongside emerging generation paradigms. 
% Early detection frameworks primarily focused on GAN-generated images, employing spatial artifact analysis through handcrafted features such as chromatic aberration~\cite{mayer2018accurate}, color~\cite{mccloskey2018detecting}, saturation~\cite{mccloskey2019detecting}, blending~\cite{li2020face}, co-occurrence~\cite{nataraj2019detecting}, and reflections~\cite{o2012exposing} or CNN classifiers~\cite{liu2020globalGramNet,wang2020cnnspot}, though these often struggled with generalization across architectures~\cite{marra2018detection,wang2020cnnspot}.
Early detection frameworks primarily focused on GAN-generated images, employing spatial artifact analysis through handcrafted features such as chromatic aberration~\cite{mayer2018accurate}, color~\cite{mccloskey2018detecting}, and saturation~\cite{mccloskey2019detecting}, though these often struggled with generalization across architectures~\cite{marra2018detection,wang2020cnnspot}.
Subsequent approaches leveraged frequency-domain inconsistencies caused by upsampling operations~\cite{frank2020leveraging,jeong2022bihpf} and model-specific fingerprints~\cite{marra2019gans,yu2019attributing,wesselkamp2022misleading} to improve cross-model robustness.
The rise of pretrained vision models introduced new paradigms, where frozen feature extractors~\cite{radford2021learning,ojha2023towards} or gradient-based representations~\cite{tan2023learning} were harnessed for forgery cues, though debates persist about the necessity of feature adaptation versus static extraction. 
Few-shot learning~\cite{tang2023m3net,tang2022learning,tang2025connecting}, is also a potential direction.
As diffusion models redefine generation quality, recent studies reveal critical performance gaps in detectors trained on GAN artifacts~\cite{corvi2023intriguing,corvi2023detection}. 
This prompts the exploration of diffusion-specific signatures~\cite{sha2023defake} via spectral analysis~\cite{corvi2023intriguing,ricker2022towards}, hybrid features~\cite{yan2025sanity}, and reconstruction errors in the pixel space~\cite{wang2023dire,ma2023exposing}.
Besides, language-guided contrastive learning~\cite{wu2023generalizable} also shows promise.
Recently, FatFormer~\cite{liu2024forgery} finds that the forgery adaptation of pre-trained models is essential for the generalizability of synthetic image detection.
LARE~\cite{2403.17465v3} shows that reconstruction error in the latent space can also benefit diffusion-generated image detection.
DRCT~\cite{chen2024drct} utilizes diffusion reconstruction to generate hard samples for contrastive learning.
However, computational efficiency and universal applicability across generative frameworks remain challenging.

%%% Local Variables:
%%% mode: LaTeX
%%% TeX-master: "../anonymous-submission-latex-2026.tex"
%%% End:

%% file: sec/3_method.tex
\section{Methodology}
\label{sec:method}

% In this paper, we present a novel diffusion-generated image detection method termed FIND, which eliminates the need for the reconstruction process used by previous methods.

% \input{sub_tex/figures/fig2_3}
\subsection{Preliminaries}
Denoising Diffusion Probabilistic Models (DDPMs)~\cite{ho2020denoising} has emerged as a significant class within the realm of generative models, revolutionizing the landscape of image generation.
The DDPMs involves two key parts: the forward and reverse diffusion processes.
In the forward process, which follows a Markov chain, Gaussian noise is incrementally added to the original image $x_0$ until degenerating it into isotropic Gaussian distribution.
Mathematically, the transition from the image $\bm{x}_{t - 1}$ at step $t-1$ to the noisy image $\bm{x}_t$ at step $t$ is defined as:
\begin{equation}
\label{eq:ddpm_forward}
 q(\bm{x}_t|\bm{x}_{t - 1})=\mathcal{N}(\bm{x}_t;\sqrt{\frac{\alpha_{t}}{\alpha_{t-1}}}\bm{x}_{t - 1},(1 - \frac{\alpha_{t}}{\alpha_{t-1}})\bm{\mathrm{I}}),
\end{equation}
where $\alpha_1, \cdots, \alpha_{T}$ are pre-defined noise schedule, and $\bm{\mathrm{I}}$ is the identity matrix. 
Leveraging the properties of the Markov process and Gaussian distributions, we can directly obtain $\bm{x}_t$ from $\bm{x}_0$:
\begin{equation}
q(\bm{x}_t|\bm{x}_0)=\mathcal{N}(\bm{x}_t;\sqrt{\alpha_t}\bm{x}_0,(1 - \alpha_t)\bm{\mathrm{I}}).
\end{equation}
The reverse diffusion process aims to reverse the corruption, which is also defined as a Markov chain, with the reverse transition probability:
\begin{equation}
\label{eq:ddpm_reverse}
p_{\theta}(\bm{x}_{t - 1}|\bm{x}_t)=\mathcal{N}(\bm{x}_{t - 1};\mu_{\theta}(\bm{x}_t,t),\Sigma_{\theta}(\bm{x}_t,t)),
\end{equation}
where $\mu_{\theta}(\bm{x}_t,t)$ and $\Sigma_{\theta}(\bm{x}_t,t)$ are functions parameterized by a neural network $\theta$. 
Instead of directly predicting $\bm{x}_{t - 1}$, the network is often trained to predict the noise $\bm{\epsilon}$ added in the forward process. 
During training, the network $\bm{\epsilon}_{\theta}$ is optimized to minimize the loss function:
\begin{align}
L_{simple}(\theta)=  \mathbb{E}_{t,\bm{x}_0,\bm{\epsilon}}&\left[\left\|\bm{\epsilon}-\bm{\epsilon}_{\theta}(\sqrt{\alpha_t}\bm{x}_0  +\sqrt{1 - \alpha_t}\bm{\epsilon},t)\right\|^2\right], 
\nonumber \\
& \bm{\epsilon} \sim\mathcal{N}(\bm{0},\bm{\mathrm{I}}).
\end{align}
% where $\bm{\epsilon}\sim\mathcal{N}(\bm{0},\bm{\mathrm{I}})$. 
%
% \input{sub_tex/figures/gmm}
% \begin{figure}[t]
% 	\centering
% 	\includegraphics[width=0.8\columnwidth]{figures/FIND_OVERALL.pdf}
% 	% \vspace{-1.5em}
%     \caption{The training framework of FIND. Gaussian noise is added to real images, and these perturbed versions are labeled as synthetic within the training process. This enables FIND to mitigate the reconstruction model dependency, allowing it to learn the core distributional differences between real and synthetic images, resulting in a superior generalization ability.}
% 	\label{fig:ci_vis}
% \end{figure}

\subsection{Hypothesis about Reconstruction Error}
\label{sec:Hypothesis}
In prior methods, \emph{e.g.}, DIRE and LaRE$^{2}$, the input images or latent features are passed through the forward and reverse processes defined in Eq.\,\ref{eq:ddpm_forward} and Eq.\,\ref{eq:ddpm_reverse} for the reconstruction error.
Despite substantial efforts to minimize the cost associated with these two processes, an additional reconstruction model is still required, along with at least one iteration of the forward and reverse processes.

We aim to eliminate the need for reconstruction error by understanding its underlying meaning. 
Considering that both the forward and reverse processes of the diffusion model involve adding Gaussian noise gradually, 
we hypothesize that the reconstruction error essentially measures how well a data distribution can be fitted by a Gaussian mixture model.
Specifically, let the distribution of real images be $(P_{real}(\bm{x})$, which can be expressed as a superposition of a large, possibly infinite, number of Gaussian distributions:
\begin{equation}
\label{eq:p_real}
P_{real}(\bm{x})=\sum_{i = 1}^{N_1}w_i\mathcal{N}(\bm{x};\bm{\mu}_i,\bm{\sigma}_i^2),
\end{equation}
where $N_1$ is a large number, $w_i$ are weights such that $\sum_{i = 1}^{N_1}w_i = 1$, and $\mathcal{N}(x;\mu_i,\sigma_i^2)$ is a Gaussian distribution with mean $\mu_i$ and variance $\sigma_i^2$.
Let the distribution of synthetic images be $P_{synthetic}(x)$. Since synthetic images are generated by a diffusion model through finite iterations, it is a superposition of finite Gaussian distributions:
\begin{equation}
P_{synthetic}(\bm{x})=\sum_{j = 1}^{N_2}v_j\mathcal{N}(\bm{x};\bm{\mu}_j,\bm{\sigma}_j^2),
\end{equation}
where $v_j$ are the weights with $\sum_{j = 1}^{N_2}v_j = 1$, and $N_2 \ll N_{1}$.
We use a Gaussian mixture distribution $Q(\bm{x})=\sum_{k = 1}^{K}u_k\mathcal{N}(\bm{x};\bm{\mu}_k,\bm{\sigma}_k^2)$ to fit the above distributions,
with minimizing the KL divergence $D_{KL}(P||Q)=\int P(\bm{x})\log\frac{P(\bm{x})}{Q(\bm{x})}d\bm{x}$.
For the synthetic image distribution $P_{synthetic}(x)$, since it is already a superposition of Gaussian distributions with relatively small numbers, because the structure of $Q(\bm{x})$ is similar to $P_{synthetic}(\bm{x})$, it is easier to find suitable parameters $u_k,\bm{\mu}_k,\bm{\sigma}_k^2$ such that $D_{KL}(P_{synthetic}||Q)$ is small. 
In contrast, for the real image distribution, since the number of Gaussian components is much larger than the number in $Q$, it is more difficult to fit, resulting in a larger reconstruction error.
Therefore,  the goodness of fit of $Q(\bm{x})$, \emph{i.e.}, the reconstruction error, can serve as a discriminative feature to distinguish real images and synthetic images.

For a binary classifier $f(\cdot)$, we expect it to capture such discriminative features to identify synthetic images.
However, the reconstruction error feature is not prominent and may interfere with other features such as visual elements.
As a result, the classifier $f(\cdot)$ may be misinformed by biases in the training dataset.
Previous methods explicitly construct the reconstruction error and feed it into $f(\cdot)$ so that $f(\cdot)$ can directly obtain the corresponding information. 
For example, in DIRE, this process can be represented as
% \begin{equation}
% f(\text{DIRE}(\bm{x})) = f(|\bm{x} - R(I(\bm{x}))|),
% \end{equation}
$f(\text{DIRE}(\bm{x})) = f(|\bm{x} - R(I(\bm{x}))|)$,
where $|\cdot|$ denotes the absolute value, $I(\cdot)$ and $R(\cdot)$ are the inversion process adding noise and reconstruction process, respectively.
Both of the processes are multiple-steps, and can be also further compressed into one step~\cite{2403.17465v3}.
Note that the noise-adding process $I(\cdot)$ does not require an additional model, and $R(\cdot)$ can be represented as a forward neural network.
This inspires us to implicitly incorporate the $R(\cdot)$ process into the classifier $f(\cdot)$ to eliminate the dependence on the reconstruction process and the reconstruction model.

Adding random noise $\bm{\epsilon} \sim \mathcal{N}(\bm{0},\bm{\sigma}_n^2)$ to real images, 
the distribution of the noisy images $P_{noisy}(\bm{x})$ is formulated as:
\begin{align}
\label{eq:p_noisy}
P_{noisy}(x) &=P_{real}(\bm{x})*\mathcal{N}(\bm{0},\bm{\sigma}_n^2) \nonumber \\
&=\sum_{i = 1}^{N_1}w_i(\mathcal{N}(\bm{x};\bm{\mu}_i,\bm{\sigma}_i^2)*\mathcal{N}(\bm{0},\bm{\sigma}_n^2)) \\
&=\sum_{i = 1}^{N_1}w_i\mathcal{N}(\bm{x};\bm{\mu}_i,\bm{\sigma}_i^2+\bm{\sigma}_n^2). \nonumber
\end{align}
Compare with $P_{real}$ in Eq.\,\ref{eq:p_real}, $P_{noisy}$ increases the variance of the Gaussian distributions, making the entire distribution smoother. 
As the variance increases, the width of the Gaussian distributions increases, and the overlapping regions between different Gaussian components become larger, thereby reducing the number of Gaussian distributions with distinct main effects.
We also provide a visual example in Fig.\,\ref{fig:gmm}. 
The left-hand figure shows the joint distribution (green line) composed of three Gaussian distributions and the fitting results (red line) using a Gaussian mixture model with two components.
In the right-hand figure, we increase the variances of the three Gaussian distributions. 
It can be seen that the joint distribution becomes smoother, and the two Gaussian distributions on the left part almost merge. 
When fitting this distribution, the mean squared error of the fitting decreases from $3.9\times 10^{-4}$ to $8\times 10^{-5}$.

From the information-theoretic perspective, 
the Fisher information matrix of each component is $\frac{w_i}{\bm{\sigma}_i^2}$ for the real image distribution $P_{real}(\bm{x})$, and $\frac{w_i}{\bm{\sigma}_i^2+\bm{\sigma}_n^2}$ for the noisy image distribution $P_{noisy}(\bm{x})$.
After adding random Gaussian noise, the Fisher information of the distribution decreases, so that this distribution is easier to be fitted by the finite Gaussian mixture distribution $Q(\bm{x})$. 

Therefore, such a noisy distribution is consistent with the synthetic image distribution in terms of the discriminative feature of the reconstruction error. 
Moreover, in other dimensions such as visual elements, since this distribution is obtained by adding random Gaussian noise to real images, it is consistent with the real image distribution.

\input{sub_tex/alg}

\input{sub_tex/figures/performance}

\subsection{Implementation of the FIND}
Based on the above analysis, we present our proposed method FIND.
Considering such noisy data is align with synthetic images on reconstruction error and consistent with real images on other dimensions, we can add this type of data to the training dataset and label them as synthetic data, prompting the classifier $f(\cdot)$ to focus more on the discriminative features.

Specifically, as shown in Fig.\,\ref{fig:ci_vis}, during the training phase, given a batch of data $B = \{(\bm{x}_1, y_1), (\bm{x}_2, y_2),\cdots\}$, where $\bm{x}_i\in [0, 255]^d$ denotes the image and label $y\in\{0, 1\}$, with $0$ representing synthetic images and $1$ representing real images.
We pick the real images in the batch $B$ and add random Gaussian noise to them as:
\begin{equation}
\bm{x}_i'= \Pi_{Img}(\bm{x}_i + \epsilon\cdot \bm{\eta}), \bm{\eta} \sim \mathcal{N}(\bm{0}, \bm{\mathrm{I}}),
\end{equation}
where $\epsilon$ is the hyper-parameter that controls the magnitude of the noise, and $\Pi_{Img}(\cdot) = clip(\cdot, 0, 255)$ is used to projects the input back to the image space, \emph{i.e.,} $[0, 255]^d$.
%
% Some examples of synthetic images, real images, and corresponding noisy images can be found in Fig.\,\ref{fig:examples}.

With the noisy images, we augment the original batch as:
\begin{equation}
B'=B \cup \{(\bm{x}_i', 0) | \bm{x}_{i} \in B \land y_{i} = 1 \}.
\end{equation}
Then we perform the normal binary classifier optimization with backpropagation to update $f(\cdot)$.
During the testing phase, we directly input the test images into the classifier $f(\cdot)$ to obtain the discrimination results, without the need of extra Gaussian noise or reconstruction process.
A rough illustration of the entire training process is provided in Alg.\,\ref{ag_proposed}.
% Note that the noisy images can be generated in parallel in practice.

%%% Local Variables:
%%% mode: LaTeX
%%% TeX-master: "../anonymous-submission-latex-2026.tex"
%%% End:

%% file: sub_tex/alg.tex
% \begin{algorithm}[!t]
% \SetAlgoLined
% \caption{Training procedure of FIND.}
% \label{ag_proposed}
% \KwIn{Training set $D$, noise hyper-parameter $\epsilon$, number of training epochs $T$.}
% \KwOut{Optimized classifier $f$.}
% Initialize parameters $\bm{\theta}$ of classifier $f$\;
% \For{epoch $= 1$ \KwTo $T$}{
%     \ForEach{batch $B = \{(\bm{x}_i, \bm{y}_i)\}$ in $D$}{
%         Initialize augmented batch $B' \leftarrow \emptyset$\;
%         \ForEach{$(\bm{x}_i, y_i)$ in $B$}{
%             Add original example $(\bm{x}_i, y_i)$ to $B'$\;
%             \If{$y_i = 1$}{
%                 Sample Gaussian noise $\bm{\eta} \sim \mathcal{N}(\bm{0}, \bm{\mathbf{I}})$\;
%                 Generate noisy image $\bm{x}'_i \leftarrow \Pi_{Img}(\bm{x}_i + \epsilon \cdot \bm{\eta})$\;
%                 Add synthetic example $(\bm{x}'_i, 0)$ to $B'$\;
%             }
%         }
%         Update $\bm{\theta}$ via backpropagation on $B'$\;
%     }
% }
% \Return{trained classifier $f$}.
% \end{algorithm}

\begin{algorithm}[!t]
\caption{Training procedure of FIND.}
\label{ag_proposed}
\begin{algorithmic}[1]
\REQUIRE Training set $D$, noise hyper-parameter $\epsilon$, number of training epochs $T$.
\STATE Initialize parameters $\bm{\theta}$ of classifier $f$\;
\FOR{epoch $= 1$ to $T$}
    \FOR{batch $B = \{(\bm{x}_i, \bm{y}_i)\}$ in $D$}
        \STATE Initialize augmented batch $B' \leftarrow \emptyset$\;
        \FOR{$(\bm{x}_i, y_i)$ in $B$}
            \STATE Add original example $(\bm{x}_i, y_i)$ to $B'$\;
            \IF{$y_i = 1$}
                \STATE Sample Gaussian noise $\bm{\eta} \sim \mathcal{N}(\bm{0}, \bm{\mathbf{I}})$\;
                \STATE Generate noisy image $\bm{x}'_i \leftarrow \Pi_{Img}(\bm{x}_i + \epsilon \cdot \bm{\eta})$\;
               \STATE  Add synthetic example $(\bm{x}'_i, 0)$ to $B'$\;
            \ENDIF
            
        \ENDFOR
        \STATE Update $\bm{\theta}$ via backpropagation on $B'$\;
    \ENDFOR
\ENDFOR
\RETURN trained classifier $f$.
\end{algorithmic}
\end{algorithm}

%%% Local Variables:
%%% mode: LaTeX
%%% TeX-master: "../main"
%%% End:

%% file: sub_tex/figures/performance.tex
\begin{figure*}[t]
\centerline{\includegraphics[width=\linewidth]{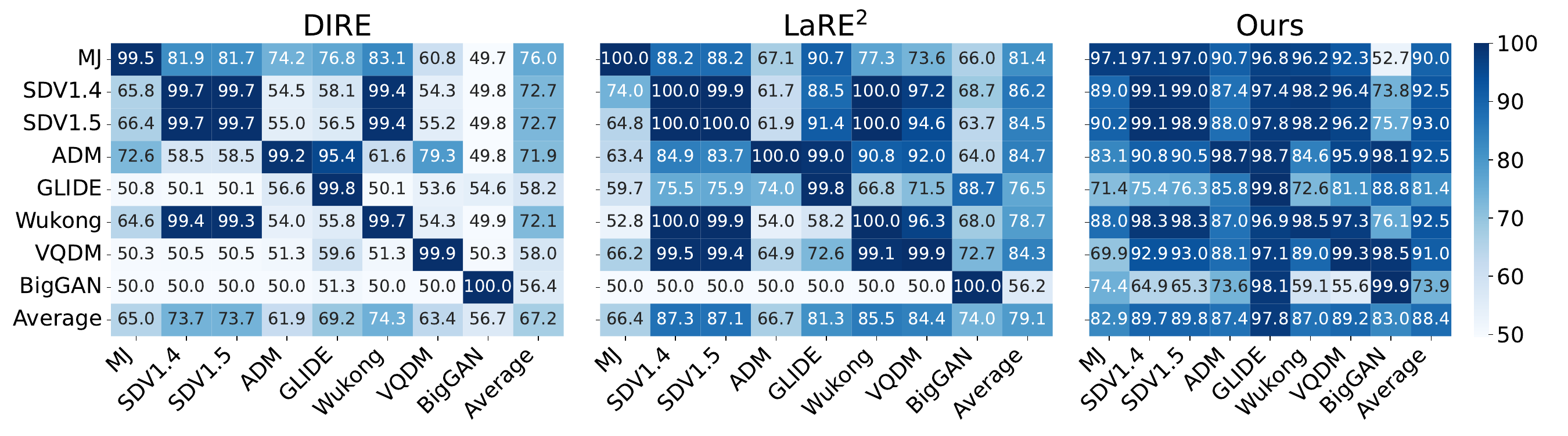}}
\caption[]{\label{fig:performance}
Cross-validation results across different training and testing subsets. 
We report the accuracy for DIRE, LaRE$^{2}$ and Ours on all 8 generators, where each row represents a corresponding training subset.
Our method shows consistent and superior performance on different training sets and test sets.
}
\end{figure*}

%%% Local Variables:
%%% mode: LaTeX
%%% TeX-master: "../../main"
%%% End:

%% file: sec/4_exp.tex
\section{Experiments}
\label{sec:experiment}
\input{sub_tex/tables/tb_main}

% In this section, we mainly evaluate our proposed FIND along with baselines on the widely-used benchmark to validate the effectiveness and efficiency of FIND. 
%
% We first introduce the experimental setups, including benchmarks, baseline methods and other experimental details.
% We first introduce the experimental setups, including benchmarks and other experimental details.
%
% Then we compare our method with competitive baselines.
% We also provide ablation studies and parameters experiments.

\subsection{Experimental Setups}
\label{sec:experimental_setup}
\textbf{Datasets and Evaluation Metric.}
All the experiments are conducted on the large-scale benchmark, GenImage~\cite{2306.08571v2}.
This benchmark contains a total of 2,681,167 images, which are partitioned into 1,331,167 real images and 1,350,000 synthetic images.
The real images are all from ImageNet~\cite{deng2009imagenet}.
By leveraging 1000 distinct labels from ImageNet for generation, it ensures a nearly equal distribution of real and generated images across each class. 
The synthetic images are generated from 8 different generative models: BigGAN~\cite{brock2018largebiggan}, GLIDE~\cite{nichol2021glide}, VQDM~\cite{nichol2021glide}, Stable Diffusion V1.4~\cite{rombach2022high},  Stable Diffusion V1.5~\cite{rombach2022high}, ADM~\cite{2105.05233v3}, Midjourney~\cite{Midjourney}, and Wukong~\cite{wukong} 
Each generator creates a similar quantity of images per class.
This dataset is organized into 8 subsets accordingly, where each subset consists of the synthetic images generated by a single generator and their corresponding real images. 
Notably, real images are not shared across subsets, and the number of generated images closely matches the number of real images in both the whole dataset and each subset.
We adhere to the official division of the dataset, with 2,581,167 images for training and the remaining 100,000 images for validation. 
We report Accuracy (ACC) as evaluation metric.

% \noindent\textbf{Baseline.}
% We compare our method with .

\noindent\textbf{Implementation Details.}
Our method is implemented based on the pytorch-image-models library~\cite{rw2019timm}. 
We adhere to the default data augmentation techniques therein, which encompass operations such as random resizing and cropping, as well as random erasing. 
We mainly utilize the ViT-B/16~\cite{dosovitskiy2020image} architecture as our backbone, taking into account its representational capabilities.
For a fair comparison, following previous work~\cite{wang2023dire,2403.17465v3}, we also tested the performance of our model using ResNet50~\cite{he2016deepresnet} as the backbone. 
Both the ViT-B/16 and ResNet50 models are initialized from the pre-trained CLIP~\cite{radford2021learning} following previous works~\cite{2403.17465v3,liu2024forgery}.
The training process is carried out on an NVIDIA A100 Tensor Core GPU. 
We conduct training for 30 epochs with a batch size of 96. 
The Stochastic Gradient Descent (SGD) optimization method is employed, and the learning rate (lr) is set to 1e-4.
This combination of optimization method and learning rate setting has been carefully tuned to ensure efficient convergence during the training process.
We set the hyper-parameter $\epsilon$ of Gaussian noise as 50.
We train 8 models separately on 8 distinct subsets. Each subset corresponds to a different generation method. 
Model selection is then based on the performance in the validation set, and the validation set shares the same generator as the training set.

\subsection{Comparison with State-of-the-Art Methods}
We compare the accuracy of our method with existing methods in Tab.\,\ref{tab:main} and Fig.\,\ref{fig:performance}.
In Tab.\,\ref{tab:main}, values in each row represent the average test accuracies of the corresponding method. 
This average is calculated when the method is trained on 8 subsets and tested on the remaining test set. 
% The last column presents the overall average of accuracy.
Compared with existing methods, our method demonstrates a clear superior performance.
Specifically, when using the ResNet50 as the backbone of the classifier, the average accuracy of our method reaches 80.13\%, which is higher than SOTA methods such as LaRE$^{2}$ (79.07\%) and FatFormer (79.04\%).
We also employ the ViT-B/16 as the backbone due to its stronger capability of representation.
Then the average accuracy is further increased by 10.3\% to reach 88.35\%. 
Our method performs stably on different test sets from different generators, and outperforms existing methods.

We further present the detailed accuracies in the form of matrix in Fig.\,\ref{fig:performance} for DIRE, LaRE$^{2}$ and our method.
For DIRE, the values of accuracy on the diagonal, where the training set and the test set are from the same generative model, are significantly higher than those in other positions.
Through means like optimization in latent space, LaRE$^{2}$ has shown improvements compared to DIRE.
However, it still falls short.  
Moreover, since both DIRE and LaRE$^{2}$ use Stable Diffusion v1.5 as their reconstruction model, these two methods achieve better results on test set from SD V1.4 and SD V1.5, yet this impedes their generalization to other models.
Our method, without the reliance on the reconstruction model, is capable of achieving a more consistent and superior accuracy performance. 

\input{sub_tex/tables/tb_time}
Subsequently, we compare the inference time costs of FIND with DIRE and LaRE$^{2}$. 
The forward processes of both DIRE and LaRE can be partitioned into a reconstruction process and a classification process. 
In Tab.\,\ref{tab:time}, we report the time costs at different stages, the overall time cost, and the frames per second (FPS). 
This table is obtained by averaging the runs on 1000 samples with the batch size of 96.
As presented in Tab.\,\ref{tab:time}, for the DIRE method, it demands a reconstruction time of 151.6 milliseconds per image. 
This leads to a total time consumption of 151.9 milliseconds, resulting in a frame rate (FPS) of 6.6. 
In the case of the LaRE$^{2}$ method, it has a reconstruction time of 50.0 milliseconds, a cumulative total time of 50.4 milliseconds, and an FPS of 19.8.
It should be noted that the time cost of the classification process is far less than that of the reconstruction process, and it can even be negligible. 
In contrast, our method completely eradicates the reconstruction process. 
As a consequence, the total inference time is drastically reduced to a mere 0.4 ms with an extremely high FPS of 2500. 
% Evidently, in terms of inference efficiency, our method significantly surpasses the state-of-the-art methods. 
% This remarkable advantage makes our method highly suitable for real-time practical applications such as real-time image analysis in surveillance systems or rapid-response content filtering in online platforms. 
This remarkable advantage makes FIND suitable for real-time practical applications. 

\input{sub_tex/tables/tb_ablation}
\subsection{More Analysis}

Subsequently, we conduct an in-depth analysis of the impact of different noise-adding methods on the results. 
We take into account three types of noise and their combinations. 
Firstly, we consider using Gaussian noise as images added to the training set. 
Secondly, we add random Gaussian noise to real images. 
Thirdly, we add random Gaussian noise to synthetic images. 
All these three types of noise-added data are labeled as synthetic images.
In Tab.\,\ref{tab:ablation}, we analyze the effects of using such noisy data, where N, R w/ N and S w/ N denote the three types of noisy data mentioned above respectively.
The row 1 serves as the baseline, without any of the specific noise-adding strategies, and the model achieves an average accuracy of 80.58\%. 
Note that this result also outperforms existing methods owing to the better backbone architecture and the appropriate data augmentation strategy.
When only the Gaussian noise as images (marked by “$\checkmark$” in the N column) is used in row 2, the average accuracy drops to 75.97\%. 
This indicates that simply adding Gaussian noise images is not beneficial for improving performance.
In row 3, when both Gaussian noise as images (N) and adding random Gaussian noise to real images (R w/ N) are utilized, the average accuracy significantly increases to 85.72\%.
This shows that the combination of these two noise-adding methods can enhance the model's generalization ability. 
The accuracy improvements are observable across various test set with different generators such as Midjourney, SD V1.4, SD V1.5.
In row 4, when only adding random Gaussian noise to synthetic images (S w/ N) is applied, the average accuracy is 74.97\%, which is relatively low. 
% This implies that this single noise-adding approach may not effectively contribute to better model performance.
As hypothesized earlier, this kind of noisy data shares the same discriminative features and other interfering features with synthetic images.
Moreover, since all of them are labeled as synthetic, they do not bring in new information. 
Instead, they may exacerbate data biases, thereby leading to lower accuracy. 
In row 5, when adding random Gaussian noise to both real and synthetic images (R w/ N and S w/ N), the average accuracy reaches 84.50\%. 
This combination also shows positive impacts on the model's performance, although slightly lower than the combination in row 3.
Both row 3 and row 5 demonstrate the contribution of add noise to real images.
Notably, in row 6, when only adding random Gaussian noise to real images (R w/ N), the model achieves the highest average accuracy of 88.35\% among all the scenarios. 
The accuracy for each test generator is also high, such as 82.89\% for Midjourney and 89.69\% for SD V1.4.
This clearly demonstrates that adding random Gaussian noise to real images is the most effective noise-adding method among the considered ones, 
which significantly enhance the model's performance in distinguishing synthetic images. 

We conduct a comprehensive analysis of the impact of the magnitude $\epsilon$ of the added noise on the results. 
We perform tests on the training subset for SDv1.5, and the corresponding results of different $\epsilon$ are depicted in Fig.\,\ref{fig:acc_vs_eps}. 
From Fig.\,\ref{fig:acc_vs_eps}, it is evident that initially as the $\epsilon$ increases, the accuracy of the model experiences a rapid upward trend.
However, once the $\epsilon$ exceeds 20, the growth rate of the accuracy slows down and begins to fluctuate.
Furthermore, we illustrate the visual effects of noisy images with different $\epsilon$ in Fig.\,\ref{fig:noise_eps}. 
The first column presents the original images, while the subsequent columns display the images with different magnitudes of added noise. 
As the noise magnitude $\epsilon$ increases, the quality of the images deteriorates rapidly. 
When an extremely large amount of noise is applied, the image may degenerate into a standard Gaussian distribution, with similar effect as the second row of Tab.\,\ref{tab:ablation}. 
Here these images fail to convey any meaningful visual information.
Consequently, through a careful consideration of these factors, we opt for a more balanced value of 50 for the noise magnitude $\epsilon$. 
This strikes a reasonable compromise between leveraging the beneficial effects of noise on model performance and maintaining an acceptable level of image integrity and visual information retention. 
By choosing this value, we aim to optimize the overall performance of our method in terms of both accuracy and the utilization of relevant visual data characteristics. 

\input{sub_tex/figures/acc_eps}

%%% Local Variables:
%%% mode: LaTeX
%%% TeX-master: "../anonymous-submission-latex-2026.tex"
%%% End:

%% file: sub_tex/tables/tb_main.tex
\begin{table*}[t]
    \resizebox{\textwidth}{!}{
\begin{tabular}{l|c|cccccccc|c}
\toprule
\multirow{2}{*}{Method} & \multirow{2}{*}{Backbone} & \multicolumn{8}{c|}{Test diffusion generators} & \multirow{2}{*}{Average}                                                                                                              \\
\cline{3-10}
                        &                           & Midjourney                                    & SD V1.4        & SD V1.5        & ADM            & GLIDE          & Wukong         & VQDM           & BigGAN         &                \\
\midrule
CNNSpot~\cite{wang2020cnnspot}                 & RN50                      & 58.20                                         & 70.30          & 70.20          & 57.00          & 57.10          & 67.70          & 56.70          & 56.60          & 61.73          \\
Spec~\cite{zhang2019detectingSPEC}                    & RN50                      & 56.70                                         & 72.40          & 72.30          & 57.90          & 65.40          & 70.30          & 61.70          & 64.30          & 65.13          \\
F3Net~\cite{qian2020thinkingF3Net}                   & RN50                      & 55.10                                         & 73.10          & 73.10          & 66.50          & 57.80          & 72.30          & 62.10          & 56.50          & 64.56          \\
GramNet~\cite{liu2020globalGramNet}                 & RN50                      & 58.10                                         & 72.80          & 72.70          & 58.70          & 65.30          & 71.30          & 57.80          & 61.20          & 64.74          \\
DIRE~\cite{wang2023dire}                    & RN50                      & 65.00                                         & 73.73          & 73.69          & 61.85          & 69.16          & 74.33          & 63.43          & 56.74          & 67.24          \\
LaRE$^{2}$~\cite{2403.17465v3}                    & RN50                      & 66.36                                         & 87.26          & 87.13          & 66.70          & 81.28          & 85.50          & 84.39          & 73.98          & 79.07          \\
%LaRE$^{2}$\dag~\cite{2403.17465v3}                    & ViT-B/16                      & 66.36                                         & 87.26          & 87.13          & 66.70          & 81.28          & 85.50          & 84.39          & 73.98          & 79.07          \\
FatFormer\dag~\cite{liu2024forgery}               & ViT-B/16                      & 76.84                                         & 78.34          & 78.66          & 78.04          & 93.14          & 76.05          & 79.06          & 72.23          & 79.04          \\
\midrule
FIND                    & RN50                      & 74.03                                         & 76.02          & 75.90          & 78.90          & 92.34          & 75.22          & 82.74          & \textbf{85.90} & 80.13          \\
FIND                    & ViT-B/16                  & \textbf{82.89}                                & \textbf{89.69} & \textbf{89.77} & \textbf{87.42} & \textbf{97.82} & \textbf{87.04} & \textbf{89.25} & 82.96          & \textbf{88.35} \\
\bottomrule
\end{tabular}
}
\caption{\label{tab:main}
Comparison of the accuracy of our method with existing methods. 
Each row in the table shows the average test accuracy of the corresponding method, calculated when trained on 8 subsets and tested on the respective test set. 
The last column gives the overall average accuracy. The symbol \dag indicates our reproduction from its source code on a CLIP-ViT-B/16 version.
The proposed FIND outperforms the SOTAs on both RN50 and ViT-B/16 backbones.
}
\end{table*}

%%% Local Variables:
%%% mode: LaTeX
%%% TeX-master: "../../main"
%%% End:

%% file: sub_tex/tables/tb_time.tex
\begin{table}
\centering
\fontsize{8.5pt}{\baselineskip}\selectfont % font size
\renewcommand\tabcolsep{3pt} % column space
\renewcommand\arraystretch{1.1} % line space 
\begin{tabular}{l|rrr|r}
\toprule
\multirow{2}{*}{Method} & \multicolumn{3}{c|}{Time per image (ms)} & \multirow{2}{*}{FPS}                           \\ \cline{2-4}
                        & \multicolumn{1}{c|}{Rec}           & \multicolumn{1}{c|}{Cls} & Total &      \\ 
% \midrule
\hline
DIRE                    & \multicolumn{1}{r|}{151.6}                    & \multicolumn{1}{r|}{0.3}        & 151.9 & 6.6  \\ 
LaRE$^{2}$                    & \multicolumn{1}{r|}{50.0}                     & \multicolumn{1}{r|}{0.4}        & 50.4  & 19.8 \\
% \midrule
\hline

Ours & \multicolumn{1}{r|}{0.0} & \multicolumn{1}{r|}{0.4} & 0.4 & 2500.0 \\ 
\bottomrule
\end{tabular}
\caption{Average inference time of different methods. 
% The results are averagely obtained with 1000 samples.
Without reconstruction, our method is significantly faster.}
\label{tab:time}
% \end{center}
\end{table}

%%% Local Variables:
%%% mode: LaTeX
%%% TeX-master: "../../main"
%%% End:

%% file: sub_tex/tables/tb_ablation.tex
\begin{table*}[!t]
    \resizebox{\textwidth}{!}{
\begin{tabular}{l|ccc|cccccccc|c}
\toprule
   & N            & R w/ N          & S w/ N          & Midjourney     & SD V1.4        & SD V1.5        & ADM            & GLIDE          & Wukong         & VQDM           & BigGAN         & Average        \\
\midrule
 1 &              &              &              & 73.97          & 83.59          & 83.54          & 76.35          & 91.45          & 80.78          & 83.62          & 71.30          & 80.58          \\
 2 & $\checkmark$ &              &              & 66.66          & 78.74          & 79.00          & 73.59          & 79.83          & 75.82          & 82.43          & 71.66          & 75.97          \\
 3 & $\checkmark$ & $\checkmark$ &              & 76.03          & 87.62          & 87.70          & 86.11          & 95.03          & 83.96          & \textbf{89.58} & 79.75          & 85.72          \\
 4 &              &              & $\checkmark$ & 64.23          & 79.29          & 79.40          & 71.18          & 76.98          & 76.09          & 82.12          & 70.45          & 74.97          \\
 5 &              & $\checkmark$ & $\checkmark$ & 75.23          & 86.10          & 86.31          & 85.48          & 96.74          & 82.99          & 87.52          & 75.61          & 84.50          \\
 6 &              & $\checkmark$ &              & \textbf{82.89} & \textbf{89.69} & \textbf{89.77} & \textbf{87.42} & \textbf{97.82} & \textbf{87.04} & 89.25          & \textbf{82.96} & \textbf{88.35} \\
\bottomrule
\end{tabular}
}
\caption{\label{tab:ablation} 
Impact of different kinds of noise.
N: using Gaussian noise as images; 
R w/ N: adding Gaussian noise to real images;
S w/ N: adding Gaussian noise to synthetic images. 
The R w/ N plays a dominant role, consistent with our hypothesis.
}
\end{table*}

%%% Local Variables:
%%% mode: LaTeX
%%% TeX-master: "../../main"
%%% End:

%% file: sub_tex/figures/acc_eps.tex
\begin{figure}[t]
% \begin{center}
% \begin{minipage}[t]{1\linewidth}

\begin{subfigure}{0.48\linewidth}
\includegraphics[width=\linewidth]{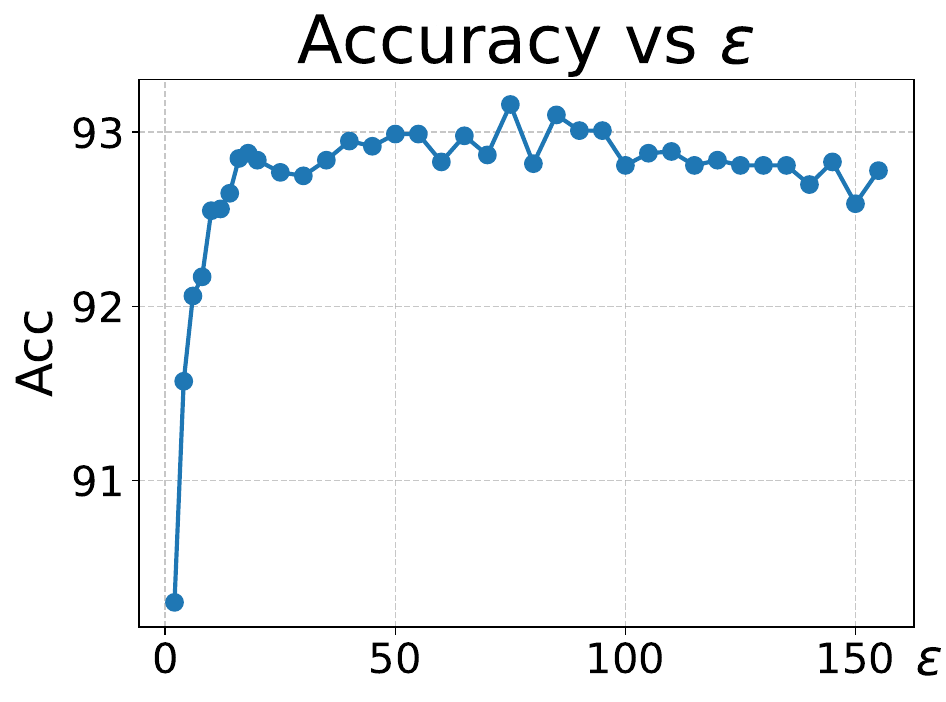}
\caption{\label{fig:acc_vs_eps} 
The effect of $\epsilon$.}
\end{subfigure}
% \end{minipage}
% \begin{minipage}[t]{0.45\linewidth}
\begin{subfigure}{0.45\linewidth}

\includegraphics[width=\linewidth]{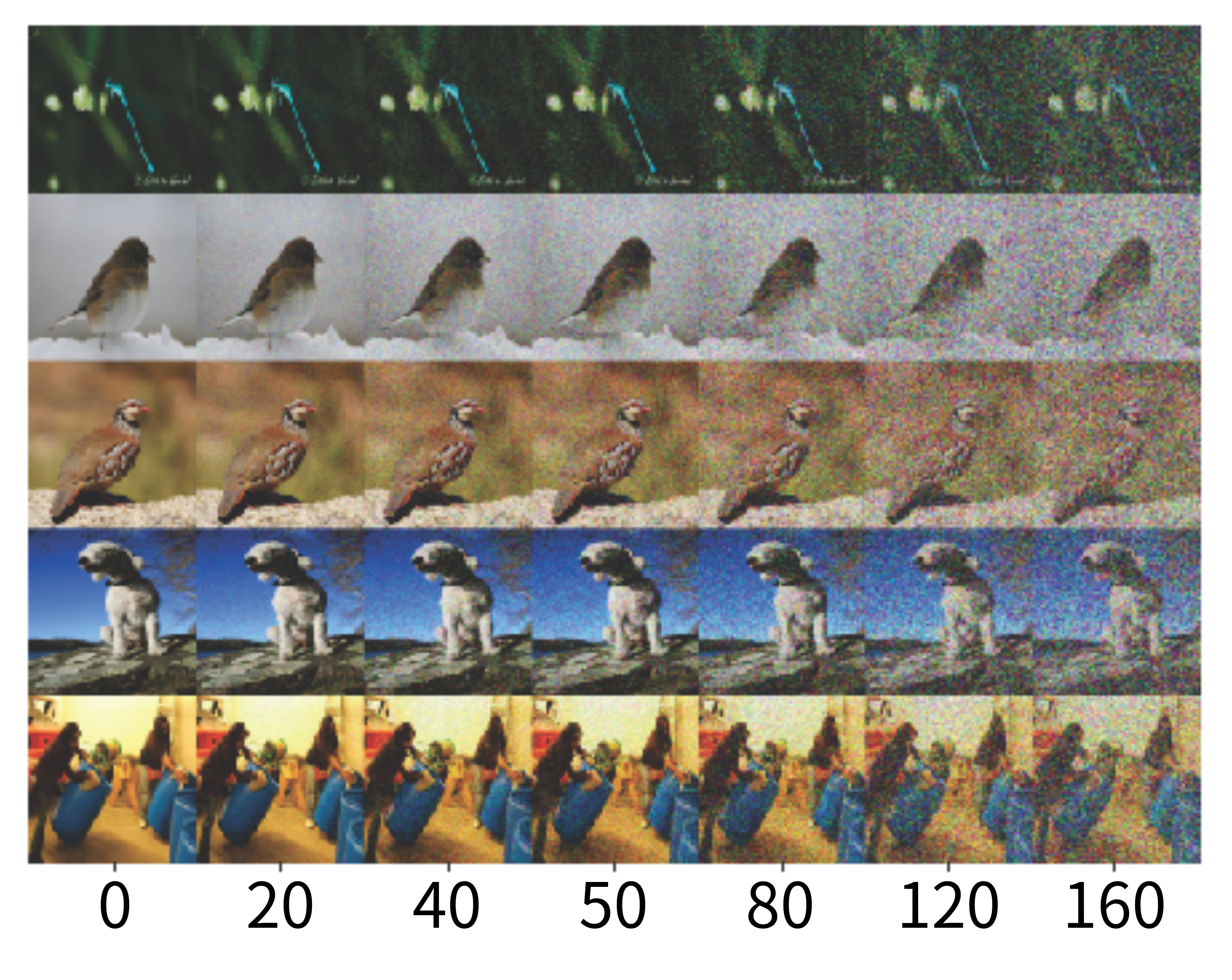}
\caption[]{\label{fig:noise_eps} Visualization with $\epsilon$.}
\end{subfigure}

% \end{minipage}

% \end{center}
\caption[]{ 
The impact of $\epsilon$.
The accuracy surges as $\epsilon$ rises initially and starts fluctuating when $\epsilon$ exceeds 20. 
The quality of the images deteriorates rapidly as $\epsilon$ increases.
We pick a balanced value of $\epsilon=50$. 
}
\end{figure}

%%% Local Variables:
%%% mode: LaTeX
%%% TeX-master: "../../main"
%%% End:

%% file: sec/5_conclusion.tex
% \section{Limitations and Broader Impact}
% \label{sec:limitation}
% While FIND demonstrates high efficiency, its design, particularly the Gaussian noise augmentation, is tailored to the statistical properties of diffusion models. 
% Consequently, its effectiveness against other generative architectures like GANs or non-Gaussian noise modeling remains an open question.
% % Additionally, like other classifiers, FIND may be susceptible to specifically crafted adversarial attacks. 
% %
% Despite these considerations, FIND's simplicity and speed significantly lower the barrier for deploying real-time detection systems, positively contributing to efforts against digital misinformation and potentially inspiring new research directions. 
% % However, the evolving nature of generative models necessitates continuous adaptation, and the ethical implications of any detection tool's errors in sensitive contexts must be carefully considered.

\section{Conclusion}
\label{sec:conclusion}
In this paper, we propose a straightforward yet efficient method named FIND for detecting images generated by diffusion models.
We first propose the hypothesis that the reconstruction error utilized by prior methods essentially reflects the fitting of Gaussian distribution.
Based on the hypothesis, we add random Gaussian noise to real images during the training of the classifier, which emphasizes the discriminative feature from reconstruction implicitly.
We evaluate our method on the large-scale benchmark, GenImage, where a higher accuracy and lower computational cost are achieved compared with existing methods.
We hope our method can serve as a new baseline for this field.

% \subsection{Limitations and Broader Impact}
% \label{sec:limitation}
% While FIND demonstrates high efficiency, its design, particularly the Gaussian noise augmentation, is tailored to the statistical properties of diffusion models. 
% Consequently, its effectiveness against other generative architectures like GANs or non-Gaussian noise modeling remains an open question.
% % Additionally, like other classifiers, FIND may be susceptible to specifically crafted adversarial attacks. 
% %
% Despite these considerations, FIND's simplicity and speed significantly lower the barrier for deploying real-time detection systems, positively contributing to efforts against digital misinformation and potentially inspiring new research directions. 
% However, the evolving nature of generative models necessitates continuous adaptation, and the ethical implications of any detection tool's errors in sensitive contexts must be carefully considered.

%%% Local Variables:
%%% mode: LaTeX
%%% TeX-master: "../main"
%%% End: